\documentclass[conference]{IEEEtran}
\IEEEoverridecommandlockouts    % to override locked commands

\usepackage{multirow}
\usepackage{lipsum}
\usepackage{amsmath}
\usepackage{amssymb}
\usepackage[ruled,vlined]{algorithm2e}
\usepackage{graphicx}
\graphicspath{{./Figures/}}
\usepackage{pgfplots}
\pgfplotsset{compat=1.18}
\usepackage{capt-of}
\usepackage{url}
\title{AROLA: A Modular Layered Architecture for Scaled Autonomous Racing}

\author{
		 	\parbox{\textwidth}{%
				 		\centering
				 		Fam Shihata$^{1}$, Mohammed Abdelazim$^{1}$, Ahmed Hussein$^{2}$% ,~\IEEEmembership{Senior Member,~IEEE}% 
				 	}%
		 	\thanks{$^{1}$German International University in Berlin (GIU Berlin), Berlin, Germany
				 		{\tt\small fam@awadlouis.com, mohammed@azab.io}}%
			\thanks{$^{2}$IAV GmbH, Berlin, Germany {\tt\small ahmed.hussein@ieee.org}}%
	}

\begin{document}
	
	\maketitle
	\thispagestyle{empty}
	\pagestyle{empty}
	
	%%%%%%%%%%%%%%%%%%%%%%%%%%%%%%%%%%%%%%%%%%%%%%%%%%%%%%%%%%%%%%%%%%
	\begin{abstract}
		Autonomous racing has advanced rapidly, particularly on scaled platforms, and software stacks must evolve accordingly. In this work, AROLA is introduced as a modular, layered software architecture in which fragmented and monolithic designs are reorganized into interchangeable layers and components connected through standardized ROS 2 interfaces. The autonomous-driving pipeline is decomposed into sensing, pre-processing, perception, localization and mapping, planning, behavior, control, and actuation, enabling rapid module replacement and objective benchmarking without reliance on custom message definitions. To support consistent performance evaluation, a Race Monitor framework is introduced as a lightweight system through which lap timing, trajectory quality, and computational load are logged in real time and standardized post-race analyses are generated. AROLA is validated in simulation and on hardware using the RoboRacer platform, including deployment at the 2025 RoboRacer IV25 competition. Together, AROLA and Race Monitor demonstrate that modularity, transparent interfaces, and systematic evaluation can accelerate development and improve reproducibility in scaled autonomous racing.
	\end{abstract}
	
	%%%%%%%%%%%%%%%%%%%%%%%%%%%%%%%%%%%%%%%%%%%%%%%%%%%%%%%%%%%%%%%%%%
	\section{Introduction}
	\label{sec:introduction}
	% !TeX root = root.tex

% Motivation and glimpse of the problem
Autonomous vehicles have grown increasingly with the rise of modern learning-based perception and planning methods, and developing and maintaining such systems has become significantly more challenging as they scale \cite{li2024complexity}. Traditional research in this domain has relied on full-scale vehicles, which introduces substantial cost, safety constraints, and reproducibility limitations. Scaled platforms, such as the RoboRacer car (formerly F1Tenth), have therefore been adopted as an accessible alternative that enables rapid experimentation, controlled validation of algorithms, and systematic testing in realistic yet predictable environments \cite{charles2025advancing}.

% Problem statement and objective
Although numerous architectures have been proposed for scaled autonomous vehicles, most existing stacks remain monolithic, tightly coupled, or tailored to specific hardware and competition rules. This lack of modularity restricts reusability, complicates benchmarking, and often results in fragmented designs that hinder comparison across research groups. To address these issues, an open, layered framework is introduced that decomposes the autonomous driving pipeline into modular components linked through standard interfaces. This structure improves system transparency, facilitates component interchangeability, and lowers the barrier to entry while preserving the expressive power required for advanced autonomous racing.

% Paper structure
The remainder of this paper introduces AROLA, a modular layered framework designed to address fragmentation and limited reusability in current stacks. Its design principles, core components, and accompanying Race Monitor evaluation suite are described. Experimental validation on both simulated and physical platforms is presented, followed by a comparison with related work and a discussion of performance metrics that illustrate AROLA’s effectiveness in both real and virtual racing environments. 

The main contributions of this work are:

\begin{enumerate}
	\item a modular, open-layered software architecture (AROLA) for scaled autonomous racing;
	\item an integrated real-time monitoring and benchmarking suite (Race Monitor); and
	\item validation in simulated and real environments.
\end{enumerate}

	%%%%%%%%%%%%%%%%%%%%%%%%%%%%%%%%%%%%%%%%%%%%%%%%%%%%%%%%%%%%%%%%%%
	\section{Related Work}
	\label{sec:related-work}
	% !TeX root = root.tex

% Related platforms and statements of the problem
Scaled autonomous driving has been widely examined, with several survey studies providing comprehensive assessments of current techniques and platforms. A central system within this field is the RoboRacer platform \cite{charles2025advancing}, whose open-source hardware and ROS-based software stack \cite{macenski2022robot} offer an effective testbed for research. Despite this, the breadth of work developed on RoboRacer has resulted in fragmentation, as many teams create architectures optimized for narrow, competition-specific scenarios. As noted by Evans et al. (2024), research on this platform has become “wide and disjointed,” complicating fair comparison and systematic evaluation \cite{evans2024unifying}. Similar concerns were raised by Serban et al. (2020), who observed that numerous autonomous vehicle prototypes were constructed in a piecemeal fashion as one-off solutions, with software architectures treated as afterthoughts rather than products of a disciplined design process \cite{serban2020standard}.

% Related software
Several notable designs have informed modular architectures for scaled autonomous racing. Kabzan et al. (2020) introduced one of the earliest full-pipeline autonomous racing systems, organized according to the sense–think–act paradigm, which influenced the decomposition used in this work \cite{kabzan2020amz}. Likewise, the TUM Autonomous Motorsport system demonstrated the effectiveness of a carefully layered architecture for high-speed racing, and its design principles contributed to the structure adopted here \cite{betz2023tum}. Further, Demeter et al. (2025) presented a two-tier software stack that separates high-performance computation from real-time control, illustrating how architectural partitioning simplifies development and execution for autonomous vehicles \cite{demeter2025autonomous}.

% Opposing related works
A contrasting direction in the literature examines end-to-end architectures. Wang et al. (2024) classified autonomous driving stacks into conventional modular systems—comprising distinct perception, localization, prediction, planning, and control subsystems—and end-to-end approaches that map raw sensor data directly to control commands \cite{wang2024moving}. The present work deliberately adopts a modular architecture to preserve interpretability and facilitate debugging, despite known issues such as error propagation across pipeline stages. End-to-end systems avoid these propagation effects but generally provide lower transparency and maintainability \cite{singh2023recent}. Recent examples include End2Race, which uses a GRU-based model to map raw LiDAR measurements directly to control commands, achieving high safety rates and strong overtaking performance \cite{qiao2025end2race}, and a deep reinforcement learning controller by Bosello et al. (2022), which demonstrated robust sim-to-real transfer using a DQN-based approach \cite{bosello2022train, mnih2015human}.

% Race monitor related works
As autonomous systems grow in complexity, systematic evaluation and benchmarking have become essential. Simulation now plays a critical role by enabling controlled testing prior to real-world deployment. However, the diversity of available frameworks complicates the selection of appropriate evaluation tools. This need has prompted the development of several benchmarking suites. Grupp’s \textit{evo} package provides standardized metrics for odometry and SLAM evaluation \cite{evo}. Evans’ f1tenth benchmarks suite supports direct comparison of classical and learning-based racing algorithms \cite{f1tenth_benchmarks}. The TUM \textit{laptime-simulation} tool further offers a quasi-steady-state vehicle-dynamics model for assessing lap time and energy consumption \cite{laptime-simulation}. Collectively, these tools emphasize the importance of reproducible, modular evaluation in advancing autonomous racing research.

These gaps motivate the need for a coherent, modular framework that standardizes interfaces, supports interchangeable components, and aligns with established robotic software practices. The following section introduces AROLA.
%, an open-layered architecture designed to address these limitations by organizing the autonomous driving pipeline into clearly defined, interoperable layers suitable for both research and competition environments.

	%%%%%%%%%%%%%%%%%%%%%%%%%%%%%%%%%%%%%%%%%%%%%%%%%%%%%%%%%%%%%%%%%%
	\section{Proposed Approach}
	\label{sec:proposed-approach}
	% !TeX root = root.tex

% Proposed approach introduction
The proposed architecture follows a modular, open-layered design intended to address limitations of monolithic and task-specific software stacks in scaled autonomous racing. AROLA (Autonomous Racing Open Layered Architecture) decomposes the autonomous driving pipeline into eight fundamental layers. Each layer is implemented as an interchangeable module connected through standardized ROS~2 interfaces to ensure transparency, extensibility, and reproducibility.

\subsection{Fundamental Components}

\subsubsection{Sensing}
It refers to the system’s ability to acquire information about the surrounding environment through onboard sensors. Modern sensing technologies enable scaled autonomous vehicles to generate large volumes of raw data that form the basis for all downstream computations \cite{ignatious2022overview}.

\subsubsection{Preprocessing}
Raw sensor data may be incomplete, noisy, or inconsistently structured. Preprocessing is therefore employed to filter, clean, and fuse these data streams before they are passed to subsequent modules. Centralizing this functionality avoids redundant operations across nodes and reduces unnecessary computational overhead.

\subsubsection{Perception}
Perception assigns semantic meaning to sensed data. While sensing captures the environment, perception interprets that information—such as identifying track boundaries, obstacles, or other agents. Keeping these layers separate improves diagnosability and supports modular development.

\subsubsection{Localization and Mapping}
Localization and mapping are treated as a coupled process, consistent with common SLAM formulations \cite{frese2006discussion}. Although they can be viewed as specialized forms of perception, they are presented as a distinct layer due to their central role in autonomous navigation and their prevalence as standalone modules in existing software stacks.

\subsubsection{Planning}
The planning layer generates feasible trajectories based on environmental perception and map information. Multiple planners may operate at different horizons or levels of granularity, enabling both global optimization and local obstacle avoidance.

\subsubsection{Behavior}
Behavior modules govern decision-making logic, typically via finite-state machines, behavior trees, or learned policies. Decoupling behavior from planning clarifies the distinction between what the vehicle should do and how the motion plan will achieve it.

\subsubsection{Control}
The control layer executes the generated trajectory while compensating for deviations arising from actuation imperfections and environmental uncertainties. This module produces low-level commands and enforces adherence to vehicle dynamics and actuator limits.

\subsubsection{Actuation}
Actuation represents the physical execution of controller commands. Steering and motor signals are applied directly to the hardware, closing the sense–think–act loop.

\subsection{Data Flow}
\begin{figure}[ht]
	\centering
	\includegraphics[width=0.4\columnwidth]{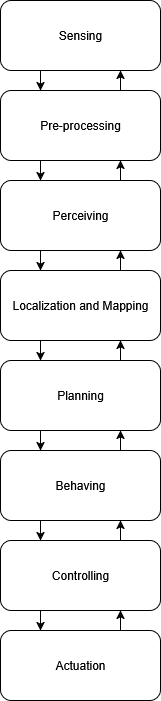}
	\caption{Overview of the complete AROLA architecture}
	\label{fig:arola-overview}
\end{figure}

AROLA is designed as an open-layered architecture that supports conventional sequential data flow while permitting optional cross-layer interactions when required. Conservative use of such interactions is recommended to preserve modularity. Figure~\ref{fig:arola-overview} illustrates the intended top-to-bottom flow.

Standard ROS~2 interfaces unify interactions between layers. Sensor outputs, perception results, planning messages, and control commands are exchanged using standard message types, such as:

\begin{itemize}
	\item $/scan$ publishing \texttt{sensor\_msgs/LaserScan}, 
	
	\item $/odom$ publishing \texttt{nav\_msgs/Odometry}, 
	
	\item $/map$ providing \texttt{nav\_msgs/OccupancyGrid} or \texttt{GetMap} services, 
	
	\item $/drive$ accepting commands as \texttt{ackermann\_msgs/} \texttt{AckermannDriveStamped} .
\end{itemize}

ROS2 tf messages provide consistent coordinate-frame alignment across nodes. Naming conventions follow standard ROS~2 practices using lowercase, descriptive identifiers. Multi-agent operation is supported through namespacing (e.g., \texttt{/ego\_racecar}, \texttt{/opp\_racecar}). The use of custom message types is deliberately minimized to enhance portability and compatibility with existing tools.

\subsection{System Integration and Deployment}
AROLA is compatible with a broad range of hardware configurations. The architecture was deployed on the RoboRacer platform using ROS~2 Humble, integrating components such as a Hokuyo LiDAR, an NVIDIA Jetson compute unit, and a VESC-based motor controller. Its hardware-agnostic design allows teams to adapt the architecture to diverse setups with minimal modification.

\begin{figure*}[ht]
	\centering
	\includegraphics[width=0.8\linewidth]{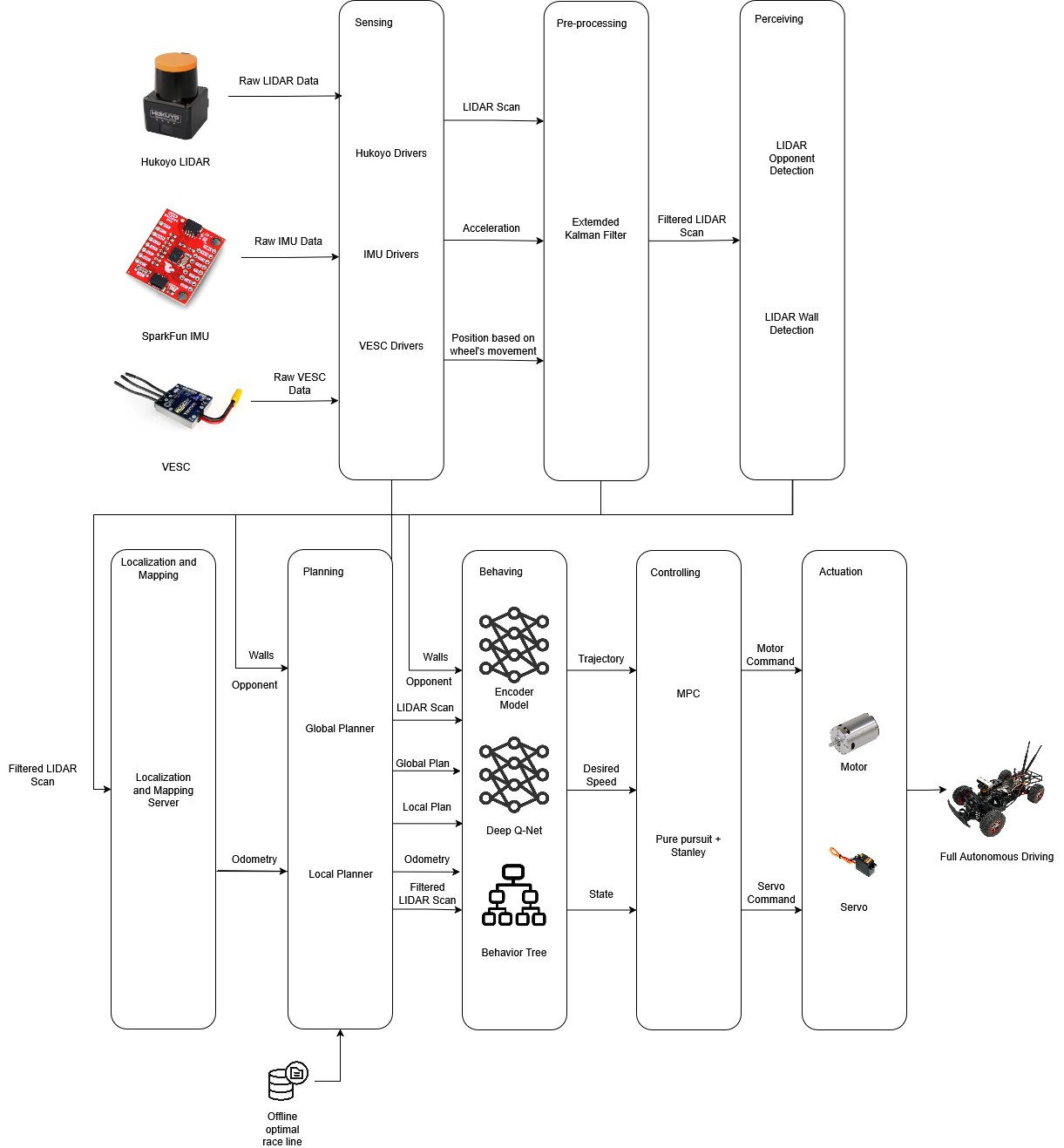}
	\caption{Example system configuration using AROLA}
	\label{fig:full-example}
\end{figure*}

The framework has also been validated across multiple simulation environments, including the RoboRacer simulator and the Forza-based racing simulator \cite{okelly2020f1tenth, baumann2024forzaeth}. These deployments demonstrate cross-platform consistency and ease of integration. A high-complexity example configuration is shown in Figure~\ref{fig:full-example}.

\subsection{Race Monitor}
To support systematic evaluation of autonomous racing performance, Race Monitor is integrated as an open, ROS~2-based monitoring framework. It provides real-time logging of lap timing, trajectory quality, and computational load, along with post-race analytics generated using \textit{evo} \cite{evo}.

Configuration is specified through a YAML file that defines enabled modules, reference trajectories, and logging parameters. Race Monitor publishes telemetry to standardized ROS~2 topics (Table~\ref{tab:race-monitor-topics}), enabling other modules—including controllers—to subscribe without introducing coupling. Reference trajectories may be provided in CSV, TUM, or KITTI formats, with start/finish zones defined interactively using RViz tools.

\begin{table}[ht]
	\renewcommand{\arraystretch}{1.2}
	\centering
	\caption{Race Monitor published ROS~2 topics}
	\label{tab:race-monitor-topics}
	\begin{tabular}{ll}
		\hline
		\textbf{Topic} & \textbf{Type} \\
		\hline
		\texttt{/race\_monitor/lap\_count} & \texttt{std\_msgs/Int32} \\
		\texttt{/race\_monitor/lap\_time} & \texttt{std\_msgs/Float32} \\
		\texttt{/race\_monitor/race\_running} & \texttt{std\_msgs/Bool} \\
		\texttt{/race\_monitor/race\_status} & \texttt{std\_msgs/String} \\
		\texttt{/race\_monitor/total\_distance} & \texttt{std\_msgs/Float32} \\
		\texttt{/race\_monitor/current\_trajectory} & \texttt{nav\_msgs/Path} \\
		\texttt{/race\_monitor/trajectory\_metrics} & \texttt{std\_msgs/String} \\
		\hline
	\end{tabular}
\end{table}

Post-race outputs include per-lap metrics, trajectory error visualizations, and performance summaries across controllers such as PID, MPC, or reinforcement-learning-based variants. The framework supports both real-time adaptation and post-hoc evaluation, enabling unbiased comparison of control strategies under identical conditions.

	%%%%%%%%%%%%%%%%%%%%%%%%%%%%%%%%%%%%%%%%%%%%%%%%%%%%%%%%%%%%%%%%%%
	\section{Experimental Work}
	\label{sec:experimental-work}
	% !TeX root = root.tex

During preparation for the RoboRacer 2025 competition, a streamlined modular software stack was deployed using AROLA. After SLAM-based map generation, a reference trajectory was produced in CSV format. A 20-minute calibration window was available for on-track testing prior to the time-trial phase. Three controllers were evaluated: (1) Gap Follower, (2) Pure Pursuit, and (3) MPC. Their performance was assessed using Race Monitor, with lap times, tracking accuracy, and computational statistics serving as evaluation metrics. A quantitative comparison of the three controllers is summarized in Table~\ref{tab:algo_comparison}.

\begin{table}[ht]
	\renewcommand{\arraystretch}{1.1}
	\centering
	\caption{Controller comparison (preliminary).}
	\label{tab:algo_comparison}
	\begin{tabular}{l r r r}
		\hline
		Metric & Gap Follower & MPC  & Pure Pursuit \\
		\hline
		Best Lap (s) & 12.60 & 10.24 & \textbf{10.10} \\
		Avg Lap (s) & 12.85 & 10.40 & \textbf{10.35} \\
		Consistency Score & -- & \textbf{0.98} & 0.92 \\
		Avg Speed (m/s) & 4.30 & 4.99 & 4.95 \\
		Control Latency (ms) & \textbf{12} & 42 & 18 \\
		CPU Load (\%) & 22 & \textbf{55} & 28 \\
		RPE Mean & -- & 4.2 & 4.3 \\
		APE Mean & -- & \textbf{0.19} & 0.22 \\
		\hline
	\end{tabular}
\end{table}

The Gap Follower exhibited strong reactivity and reliable obstacle avoidance but produced the slowest lap times. MPC and Pure Pursuit showed similarly low pose-tracking errors (APE and RPE), with MPC producing lower jerk but significantly higher computational load and control latency. Pure Pursuit generated the fastest lap times and was therefore selected for further use. After approximately 5 laps, declining lap-time consistency was observed. Race Monitor’s error-mapped trajectory visualization enabled identification and correction of suboptimal reference-path segments, followed by parameter tuning. The same minimal AROLA configuration was validated in simulation using the official RoboRacer minimal-software example.\footnote{\url{https://github.com/GIU-F1Tenth/minimal_sofware_example}}

\subsection{RoboRacer IV25 Competition Results}
The finalized configuration achieved a third-place finish with a lap time of $10.1~s$, within $0.02~s$ of second place and $1.24~s$ of first.

\subsection{Evaluation of an LQR Controller Using Race Monitor}

An external map (“Berlin Map”) was used post-competition to evaluate an LQR controller and to extend Race Monitor’s capabilities. Race Monitor logged lap statistics, tracking performance, and controller behavior during an evaluation run. The LQR controller completed seven laps without failures. The average lap time was $19.75~s$, with a best lap of $19.53~s$. Lap-time consistency was high, with a coefficient of variation of $0.15$ and a consistency score of $0.99$. Average speed reached $3.40~m/s$, with a peak of $3.79~m/s$. A summary of lap-time and tracking metrics is provided in Table~\ref{tab:race_monitor_summary}.

\begin{table}[ht]
	\renewcommand{\arraystretch}{1.1}
	\centering
	\caption{Race Monitor summary results for LQR controller (exp\_011).}
	\label{tab:race_monitor_summary}
	\begin{tabular}{l r}
		\hline
		Metric & Value \\
		\hline
		Total Laps & 7 \\
		Best Lap Time (s) & 19.53 \\
		Average Lap Time (s) & 19.75 \\
		Worst Lap Time (s) & 19.94 \\
		Lap Time Std (s) & 0.15 \\
		Consistency Score & 0.99 \\
		Average Speed (m/s) & 3.40 \\
		Max Speed (m/s) & 3.79 \\
		APE Mean & 4.044 \\
		RPE Mean & 4.42 \\
		Controller & LQR \\
		Experiment ID & exp\_011 \\
		\hline
	\end{tabular}
\end{table}

\begin{figure}[ht]
	\centering
	\includegraphics[width=0.7\linewidth]{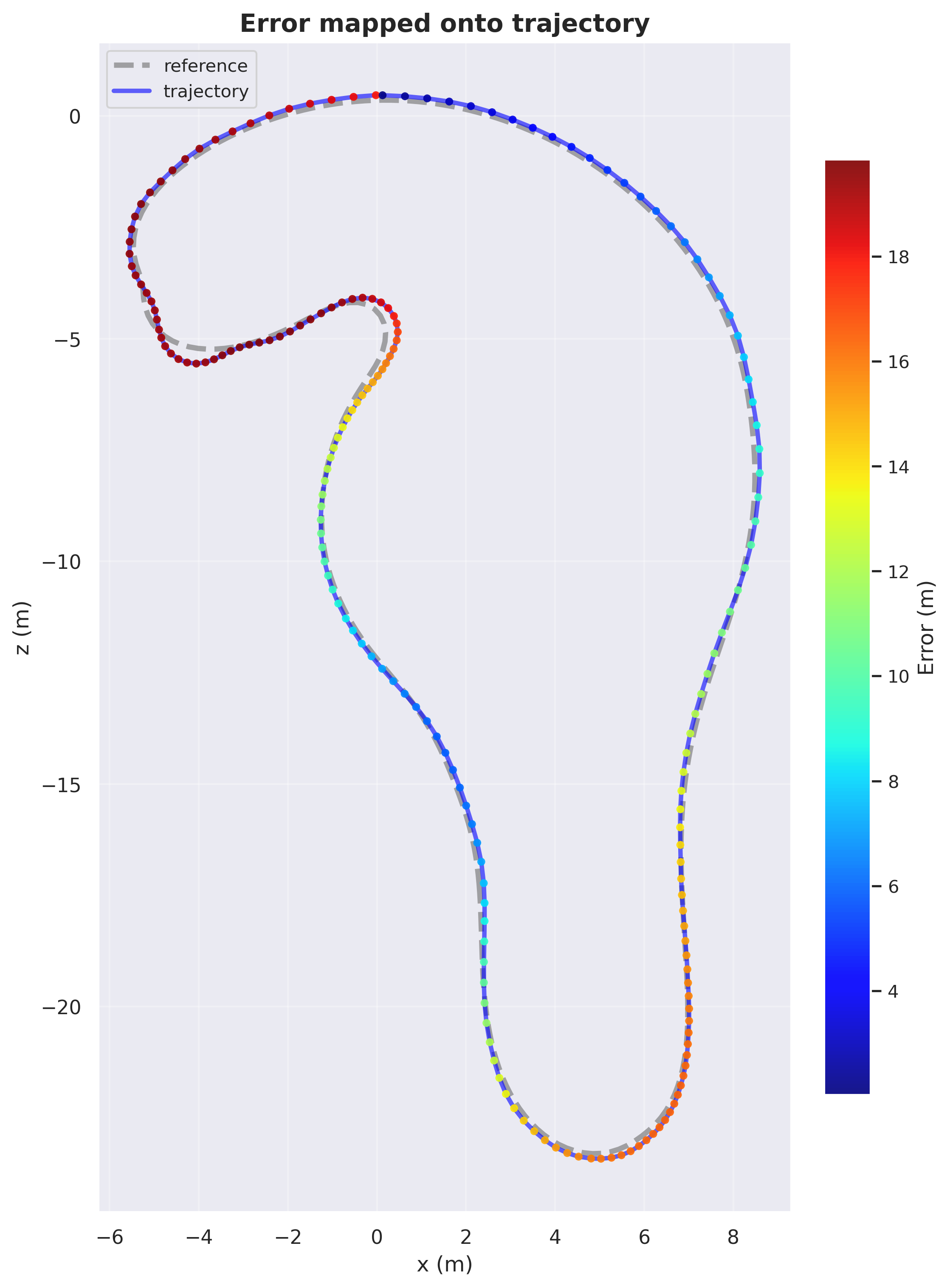}
	\caption{Error-mapped trajectory showing deviations along the LQR controller path.}
	\label{fig:best_lap_error_mapped_trajectory}
\end{figure}

Trajectory evaluation showed a low Relative Pose Error (RPE) of $4.42$. The aggregate Absolute Pose Error (APE) was higher due to a scaling misalignment at the reference trajectory’s origin rather than controller inaccuracy. Localized tracking accuracy remained stable across all laps. Figure~\ref{fig:best_lap_error_mapped_trajectory} shows the error-mapped trajectory used during parameter refinement.

\begin{figure}[ht]
	\centering
	\begin{tikzpicture}
		\begin{axis}[
			width=\linewidth,
			height=4cm,
			xlabel={Lap},
			ylabel={Time (s)},
			ymin=19.0, ymax=21.0,
			xtick={1,2,3,4,5,6,7},
			ytick distance=0.5,
			grid=both,
			title={Lap Time Trend}
			]
			\addplot+[mark=*] coordinates {
				(1,19.67) (2,19.73) (3,19.94) (4,19.84)
				(5,19.53) (6,19.58) (7,19.94)
			};
			\addplot[dashed,gray] coordinates {(1,19.75) (7,19.75)};
			\legend{Lap Time, Avg}
		\end{axis}
	\end{tikzpicture}
	\caption{Lap-time trend for the LQR controller.}
	\label{fig:lap_time_plot}
\end{figure}
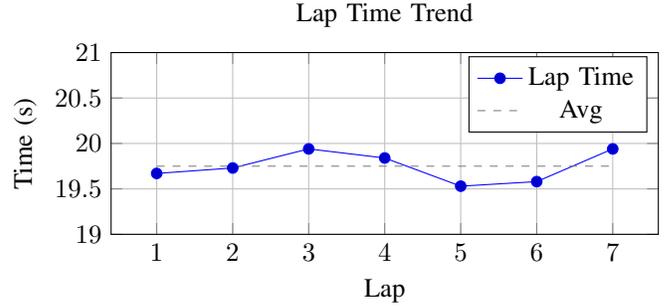

\begin{figure}[!h]
	\centering
	\includegraphics[width=0.7\linewidth]{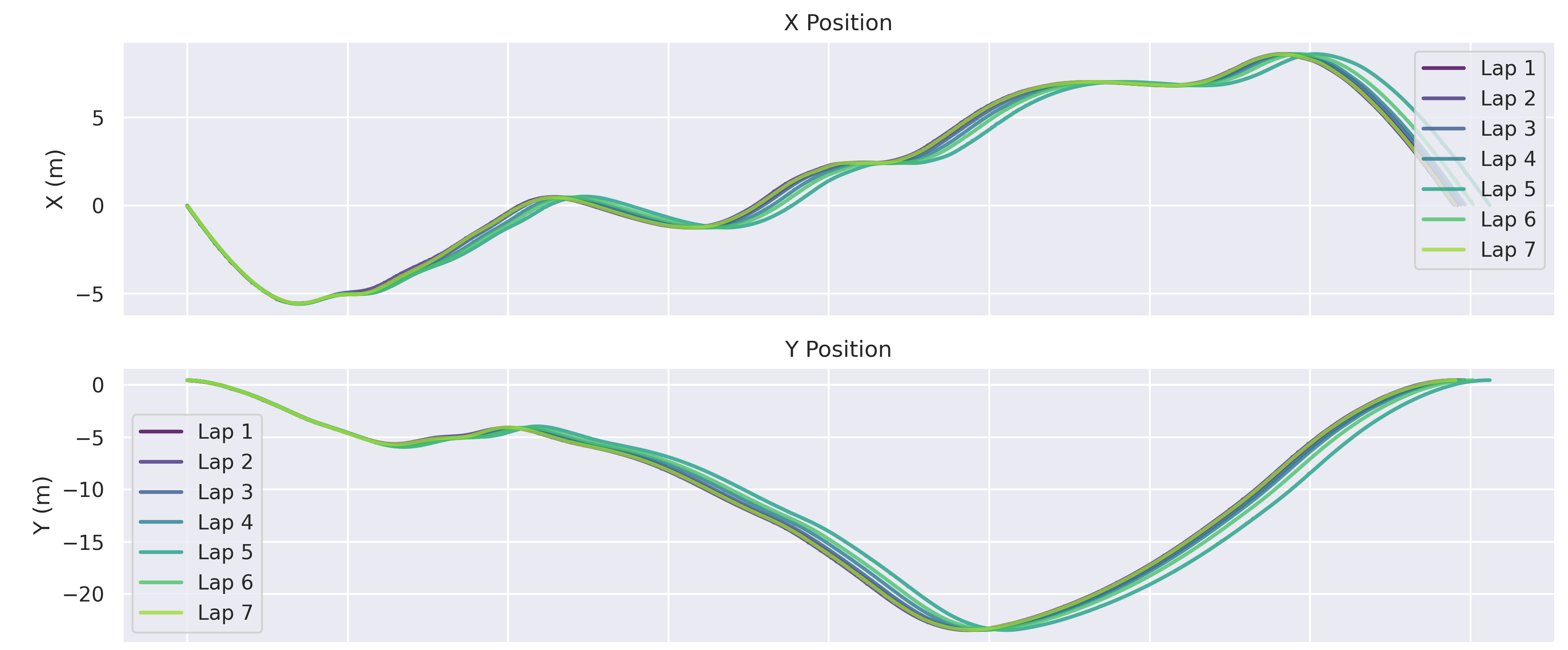}
	\caption{Per-axis translational error (XY) compared to the reference trajectory for the LQR controller (exp\_011). Z remains constant.}
	\label{fig:lqr_xyz}
\end{figure}

\begin{figure}[!h]
	\centering
	\includegraphics[width=0.7\linewidth]{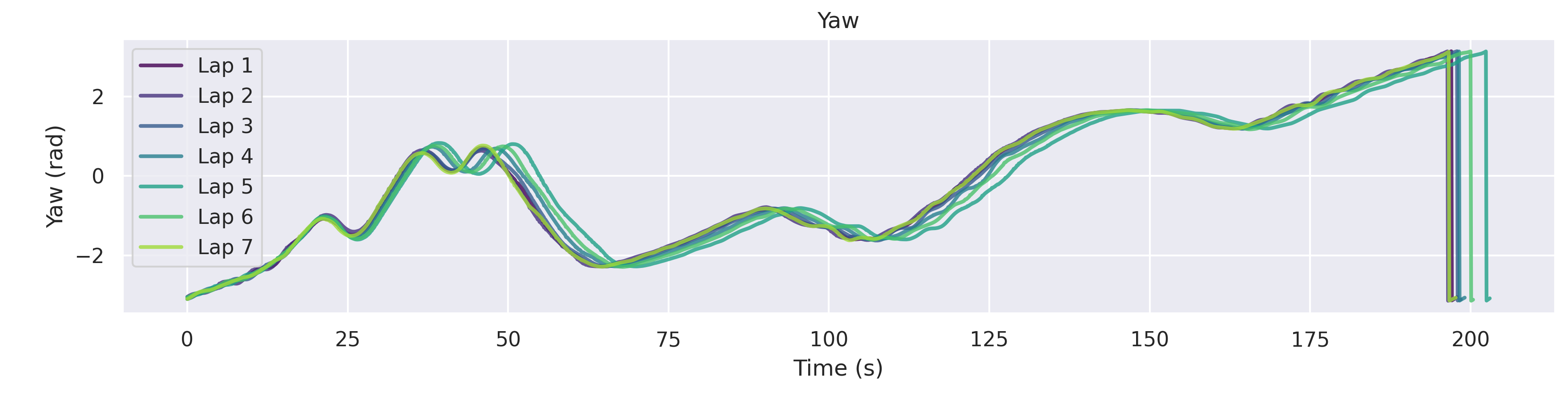}
	\caption{Yaw error distribution compared to the reference trajectory for the LQR controller (exp\_011). Roll and pitch remain constant.}
	\label{fig:lqr_rpy}
\end{figure}

Per-lap timing results for the seven recorded laps are shown in Table~\ref{tab:lap_times}, with the lap-time trend illustrated in Fig.~\ref{fig:lap_time_plot}, per-axis translational errors shown in Fig.~\ref{fig:lqr_xyz}, and orientation (yaw) errors shown in Fig.~\ref{fig:lqr_rpy}.

\begin{table}[ht]
	\renewcommand{\arraystretch}{1.1}
	\centering
	\caption{Per-lap times for LQR controller (exp\_011).}
	\label{tab:lap_times}
	\begin{tabular}{c r}
		\hline
		Lap & Time (s) \\
		\hline
		1 & 19.67 \\
		2 & 19.73 \\
		3 & 19.94 \\
		4 & 19.84 \\
		5 & \textbf{19.53} \\
		6 & 19.58 \\
		7 & 19.94 \\
		\hline
	\end{tabular}
\end{table}

%\begin{figure}[!h]
%	\centering
%	\includegraphics[width=0.7\linewidth]{Figures/Trajectory_Error_Distribution.png}
%	\caption{Trajectory error distribution for the LQR controller (exp\_011).}
%	\label{fig:lqr_error_distribution}
%\end{figure}

	%%%%%%%%%%%%%%%%%%%%%%%%%%%%%%%%%%%%%%%%%%%%%%%%%%%%%%%%%%%%%%%%%%
	\section{Summary and Conclusions}
	\label{sec:conclusion}
	% !TeX root = root.tex

AROLA is introduced as a modular software architecture aimed at reducing development effort and improving how components are reused across scaled autonomous vehicle projects. Built on the RoboRacer platform, it breaks the driving stack into clear layers—sensing, preprocessing, perception, localization or mapping, planning, behavior, control, and actuation. Each layer relies on standard ROS 2 interfaces, which keeps the system easy to extend and makes it possible to swap components without reworking the whole stack.

Race Monitor was developed alongside AROLA to track performance during both simulation and real-world runs. It records system behavior in real time and provides a consistent way to compare and tune controllers. This tool was used throughout testing and contributed to the setup that placed third at the 2025 RoboRacer IV25 event. The experiments show that AROLA’s structure makes it easier to evaluate different controllers, adjust parameters efficiently, and repeat results reliably.

The installation and deployment process is kept straightforward, with a focus on simple configuration and compatibility with different sensor and compute setups. Overall, AROLA offers a practical foundation for quick iteration, fair performance evaluation, and dependable scaling in autonomous racing research.

	%%%%%%%%%%%%%%%%%%%%%%%%%%%%%%%%%%%%%%%%%%%%%%%%%%%%%%%%%%%%%%%%%%
	\section{Limitations}
	\label{sec:limitations}
	% !TeX root = root.tex

While the proposed architecture demonstrates strong practicality and successful deployment in both simulation and hardware racing settings, its evaluation remains limited in scope. The experimental results focus primarily on a small set of controllers, and do not yet explore the broader range of perception, localization, and planning modules that the architecture is designed to support. As with most modular autonomy stacks, the system remains susceptible to error propagation across pipeline stages, where inaccuracies in perception or localization can degrade downstream planning and control. Although the architecture is designed to be hardware-agnostic and extensible, the empirical validation is centered on a specific platform and sensor configuration. Moreover, race performance metrics emphasize lap time and trajectory error while omitting other important racing considerations such as safety margins, multi-agent interactions, and robustness under disturbances.
	
	%%%%%%%%%%%%%%%%%%%%%%%%%%%%%%%%%%%%%%%%%%%%%%%%%%%%%%%%%%%%%%%%%%
	\section{Future Work}
	\label{sec:future-work}
	% !TeX root = root.tex

Further efforts will target improved tooling for trajectory, map, and coordinate-frame alignment to reduce evaluation artifacts and enhance reproducibility. Extending validation across diverse hardware platforms, sensors, simulators, and tracks will strengthen claims of portability and generality. Additionally, future works will focus on creating a watchdog system to detect misalignments with the architecture and aid in failure prevention. Moreover, Race Monitor will be extended into a web-based user interface that provides a global view of the system state and supports richer evaluation, including new metrics for benchmarking perception, localization, and planning modules alongside control performance. Finally, integrating learning-based or end-to-end components within the modular framework, while preserving standardized interfaces, represents a promising direction to balance interpretability, flexibility, and performance in autonomous racing systems. Therefore, the architecture will be extended to allow partial end-to-end components.
	
	%%%%%%%%%%%%%%%%%%%%%%%%%%%%%%%%%%%%%%%%%%%%%%%%%%%%%%%%%%%%%%%%%%
	%\addtolength{\textheight}{-12cm}
	%\vspace{10mm}
	\bibliographystyle{IEEEtran}
	% Your .bib file here
	\bibliography{root,introduction,related-work,proposed-approach,experimental-work,summary,limitations-and-future-work}

\begin{thebibliography}{10}
\providecommand{\url}[1]{#1}
\csname url@samestyle\endcsname
\providecommand{\newblock}{\relax}
\providecommand{\bibinfo}[2]{#2}
\providecommand{\BIBentrySTDinterwordspacing}{\spaceskip=0pt\relax}
\providecommand{\BIBentryALTinterwordstretchfactor}{4}
\providecommand{\BIBentryALTinterwordspacing}{\spaceskip=\fontdimen2\font plus
\BIBentryALTinterwordstretchfactor\fontdimen3\font minus
  \fontdimen4\font\relax}
\providecommand{\BIBforeignlanguage}[2]{{%
\expandafter\ifx\csname l@#1\endcsname\relax
\typeout{** WARNING: IEEEtran.bst: No hyphenation pattern has been}%
\typeout{** loaded for the language `#1'. Using the pattern for}%
\typeout{** the default language instead.}%
\else
\language=\csname l@#1\endcsname
\fi
#2}}
\providecommand{\BIBdecl}{\relax}
\BIBdecl

\bibitem{li2024complexity}
J.~Li, R.~Zong, Y.~Wang, and W.~Deng, ``Complexity evaluation for urban
  intersection scenarios in autonomous driving tests: Method and validation,''
  \emph{Applied Sciences}, vol.~14, no.~22, p. 10451, 2024.

\bibitem{charles2025advancing}
I.~Charles, H.~Maghsoumi, and Y.~Fallah, ``Advancing autonomous racing: A
  comprehensive survey of the roboracer (f1tenth) platform,'' \emph{arXiv
  preprint arXiv:2506.15899}, 2025.

\bibitem{macenski2022robot}
S.~Macenski, T.~Foote, B.~Gerkey, C.~Lalancette, and W.~Woodall, ``Robot
  operating system 2: Design, architecture, and uses in the wild,''
  \emph{Science robotics}, vol.~7, no.~66, p. eabm6074, 2022.

\bibitem{evans2024unifying}
B.~D. Evans, R.~Trumpp, M.~Caccamo, F.~Jahncke, J.~Betz, H.~W. Jordaan, and
  H.~A. Engelbrecht, ``Unifying f1tenth autonomous racing: Survey, methods and
  benchmarks,'' \emph{arXiv preprint arXiv:2402.18558}, 2024.

\bibitem{serban2020standard}
A.~Serban, E.~Poll, and J.~Visser, ``A standard driven software architecture
  for fully autonomous vehicles,'' \emph{Journal of Automotive Software
  Engineering}, vol.~1, no.~1, pp. 20--33, 2020.

\bibitem{kabzan2020amz}
J.~Kabzan, M.~I. Valls, V.~J. Reijgwart, H.~F. Hendrikx, C.~Ehmke, M.~Prajapat,
  A.~B{\"u}hler, N.~Gosala, M.~Gupta, R.~Sivanesan \emph{et~al.}, ``Amz
  driverless: The full autonomous racing system,'' \emph{Journal of Field
  Robotics}, vol.~37, no.~7, pp. 1267--1294, 2020.

\bibitem{betz2023tum}
J.~Betz, T.~Betz, F.~Fent, M.~Geisslinger, A.~Heilmeier, L.~Hermansdorfer,
  T.~Herrmann, S.~Huch, P.~Karle, M.~Lienkamp \emph{et~al.}, ``Tum autonomous
  motorsport: An autonomous racing software for the indy autonomous
  challenge,'' \emph{Journal of Field Robotics}, vol.~40, no.~4, pp. 783--809,
  2023.

\bibitem{demeter2025autonomous}
Z.~Demeter, L.~Pusk{\'a}s, B.~Kov{\'a}cs, {\'A}.~Matkovics, M.~N{\'a}das,
  B.~Tuba, Z.~Farkas, {\'A}.~Bog{\'a}r-N{\'e}meth, and G.~B{\'a}ri, ``The
  autonomous software stack of the fred-003c: The development that led to
  full-scale autonomous racing,'' \emph{arXiv preprint arXiv:2504.18439}, 2025.

\bibitem{wang2024moving}
X.~Wang, M.~A. Maleki, M.~W. Azhar, and P.~Trancoso, ``Moving forward: A review
  of autonomous driving software and hardware systems,'' \emph{arXiv preprint
  arXiv:2411.10291}, 2024.

\bibitem{singh2023recent}
P.~Singh~Chib and P.~Singh, ``Recent advancements in end-to-end autonomous
  driving using deep learning: A survey,'' \emph{arXiv e-prints}, pp.
  arXiv--2307, 2023.

\bibitem{qiao2025end2race}
Z.~Qiao, H.~Li, Z.~Cao, and H.~X. Liu, ``End2race: Efficient end-to-end
  imitation learning for real-time f1tenth racing,'' \emph{arXiv preprint
  arXiv:2509.16894}, 2025.

\bibitem{bosello2022train}
M.~Bosello, R.~Tse, and G.~Pau, ``Train in austria, race in montecarlo:
  Generalized rl for cross-track f1 tenth lidar-based races,'' in \emph{2022
  IEEE 19th Annual Consumer Communications \& Networking Conference
  (CCNC)}.\hskip 1em plus 0.5em minus 0.4em\relax IEEE, 2022, pp. 290--298.

\bibitem{mnih2015human}
V.~Mnih, K.~Kavukcuoglu, D.~Silver, A.~A. Rusu, J.~Veness, M.~G. Bellemare,
  A.~Graves, M.~Riedmiller, A.~K. Fidjeland, G.~Ostrovski \emph{et~al.},
  ``Human-level control through deep reinforcement learning,'' \emph{nature},
  vol. 518, no. 7540, pp. 529--533, 2015.

\bibitem{evo}
\BIBentryALTinterwordspacing
M.~Grupp, ``{evo}: Python package for the evaluation of odometry and slam,''
  2017, gitHub repository. Accessed 2017-01-01. [Online]. Available:
  \url{https://github.com/MichaelGrupp/evo}
\BIBentrySTDinterwordspacing

\bibitem{f1tenth_benchmarks}
B.~D. Evans, ``f1tenth\_benchmarks: Implementations of common f1tenth
  autonomous racing algorithms with benchmark results,''
  \url{https://github.com/BDEvan5/f1tenth_benchmarks}, 2023, gitHub Repository.

\bibitem{laptime-simulation}
T.~U. of~Munich Fastwheel~Motorsports, ``laptime-simulation: Quasi-steady-state
  lap time simulation,'' \url{https://github.com/TUMFTM/laptime-simulation},
  2020, gitHub Repository.

\bibitem{ignatious2022overview}
H.~A. Ignatious, M.~Khan \emph{et~al.}, ``An overview of sensors in autonomous
  vehicles,'' \emph{Procedia Computer Science}, vol. 198, pp. 736--741, 2022.

\bibitem{frese2006discussion}
U.~Frese, ``A discussion of simultaneous localization and mapping,''
  \emph{Autonomous Robots}, vol.~20, no.~1, pp. 25--42, 2006.

\bibitem{okelly2020f1tenth}
M.~O’Kelly, H.~Zheng, D.~Karthik, and R.~Mangharam, ``F1tenth: An open-source
  evaluation environment for continuous control and reinforcement learning,''
  in \emph{NeurIPS 2019 Competition and Demonstration Track}.\hskip 1em plus
  0.5em minus 0.4em\relax PMLR, 2020, pp. 77--89.

\bibitem{baumann2024forzaeth}
N.~Baumann, E.~Ghignone, J.~K{\"u}hne, N.~Bastuck, J.~Becker, N.~Imholz,
  T.~Kr{\"a}nzlin, T.~Y. Lim, M.~L{\"o}tscher, L.~Schwarzenbach \emph{et~al.},
  ``Forzaeth race stack—scaled autonomous head-to-head racing on fully
  commercial off-the-shelf hardware,'' \emph{Journal of Field Robotics}, 2024.

\end{thebibliography}
	
\end{document}